\documentclass{article}
\usepackage{color}
\usepackage[colorinlistoftodos,prependcaption, color=red!25]{todonotes}
\usepackage{natbib}
\emergencystretch=\maxdimen
\usepackage{authblk}

\begin{document}

\makeatletter
\renewcommand\AB@affilsepx{, \protect\Affilfont}
\makeatother

\title{How we do things with words: Analyzing text as social and cultural data}
\author[1,2]{Dong Nguyen}
\author[1,3]{Maria Liakata}
\author[4]{Simon DeDeo}
\author[5]{Jacob Eisenstein}
\author[6]{David Mimno}
\author[1,7]{Rebekah Tromble}
\author[8]{Jane Winters}
\affil[1]{Alan Turing Institute (UK)}
\affil[2]{Utrecht University (NL)}
\affil[3]{University of Warwick (UK)}
\affil[4]{Carnegie Mellon University (USA)}
\affil[5]{Georgia Institute of Technology (USA)}
\affil[6]{Cornell University (USA)}
\affil[7]{Leiden University (NL)}
\affil[8]{University of London (UK)}
\setcounter{Maxaffil}{0}
\renewcommand\Affilfont{\itshape\small}

\date{}
\maketitle{}
\vspace{-3.3em}
\section{Introduction}
In June 2015, the operators of the online discussion site Reddit banned several communities under new anti-harassment rules. \cite{chandrasekharan2017you} used this opportunity to combine rich online data with computational methods to study a current question: Does eliminating these ``echo chambers'' diminish the amount of hate speech overall? Exciting opportunities like these, at the intersection of ``thick'' cultural and societal questions on the one hand, and the computational analysis of rich textual data on larger-than-human scales on the other, are becoming increasingly common.

Indeed, computational analysis is opening new possibilities for exploring challenging questions at the heart of some of the most pressing contemporary cultural and social issues. While a human reader is better equipped to make logical inferences, resolve ambiguities, and apply cultural knowledge than a computer, human time and attention are limited. Moreover, many patterns are not obvious in any specific context, but only stand out in the aggregate. For example, in a landmark study, \cite{doi:10.1080/01621459.1963.10500849} analyzed the authorship of \textit{The Federalist Papers}  using a statistical text analysis by focusing on style, based on the distribution of function words, rather than content. As another example, \cite{long2016literary} studied what defines English haiku and showed how computational analysis and close reading can complement each other. Computational approaches are valuable precisely because they help us identify patterns that would not otherwise be discernible.

Yet these approaches are not a panacea. Examining thick social and cultural questions using computational text analysis carries significant challenges. For one, texts are culturally and socially situated. They reflect the ideas, values and beliefs of both their authors and their target audiences, and such subtleties of meaning and interpretation are difficult to incorporate in computational approaches. For another, many of the social and cultural concepts we seek to examine are highly contested --- hate speech is just one such example. Choices regarding how to operationalize and analyze these concepts can raise serious concerns about conceptual validity and may lead to shallow or obvious conclusions, rather than findings that reflect the depth of the questions we seek to address. 

These are just a small sample of the many opportunities and challenges faced in computational analyses of textual data. New possibilities and frustrating obstacles emerge at every stage of research, from identification of the research question to interpretation of the results. 
In this article, we take the reader through a typical research process that involves measuring social or cultural concepts using computational methods, discussing both the opportunities and complications that often arise. In the Reddit case, for example, hate speech is measured, however imperfectly, by the presence of particular words semi-automatically extracted from a machine learning algorithm. Operationalizations are never perfect translations, and are often refined over the course of an investigation, but they are crucial.

We begin our exploration with the identification of research questions, proceed through data selection, conceptualization, and operationalization, and end with analysis and the interpretation of results. 
The research process sounds more or less linear this way, but each of these phases overlaps, and in some instances turns back upon itself. 
The analysis phase, for example, often feeds back into the original research questions,  which may continue to evolve for much of the project. At each stage, our discussion is critically informed by insights from the humanities and social sciences, fields that have focused on, and worked to tackle, the challenges of textual analysis---albeit at smaller scales---since their inception.

In describing our experiences with computational text analysis, we hope to achieve three primary goals. First, we aim to shed light on thorny issues not always at the forefront of discussions about computational text analysis methods. Second, we hope to provide a set of best practices for working with thick social and cultural concepts. Our guidance is based on our own experiences and is therefore inherently imperfect. Still, given our diversity of disciplinary backgrounds and research practices, we hope to capture a range of ideas and identify commonalities that will resonate for many. And this leads to our final goal: to help promote  interdisciplinary collaborations. Interdisciplinary insights and partnerships are essential for realizing the full potential of any computational text analysis that involves social and cultural concepts, and the more we are able to bridge these divides, the more fruitful we believe our work will be.

 \paragraph*{Context and intended audience}
 This article is a result of a workshop organized in 2017 by Maria Liakata and Dong Nguyen at the Alan Turing Institute on ``Bridging disciplines in analysing text as social and cultural data''. The primary audience of this article are most likely readers who are interested in doing interdisciplinary text analysis, with a computer science, humanities, or social sciences background. We expect that this text may be especially useful for courses on text as data. 

\section{Research questions}
\label{sec:develop_research_questions}
We typically start by identifying the questions we wish to explore. 
Can text analysis provide a new perspective on a ``big question'' that has been attracting interest for years? 
Or can we raise new questions that have only recently emerged, for example about social media? 
For social scientists working in computational analysis, the questions are often grounded in theory, asking: How can we explain what we observe?
These questions are also influenced by the availability and accessibility of  data sources. For example, the choice to work with data from a particular social media platform may be partly determined by the fact that it is freely available, and this will in turn shape the kinds of questions that can be asked.   A key output of this phase are the concepts to measure, for example: influence; copying and reproduction; the creation of patterns of language use; hate speech. 

Computational analysis of text motivated by these questions is  \textbf{insight driven}: we aim to describe a phenomenon or explain how it came about. For example, what can we learn about how and why hate speech is used or how this changes over time? Is hate speech one thing, or does it comprise multiple forms of expression? Is there a clear boundary between hate speech and other types of speech, and what features make it more or less ambiguous? 
In these cases, it is critical to communicate  high-level patterns 
in terms that are recognizable.

This contrasts with much of the work in computational text analysis, which tends to focus on  automating tasks that humans perform inefficiently. These tasks range from core linguistically motivated tasks that constitute the backbone of natural language processing, such as part-of-speech tagging and parsing, to filtering spam and detecting sentiment. 
Many tasks are motivated by applications, for example to  automatically block online trolls.  Success, then, is often measured by performance, and communicating why a certain prediction was made---for example, why a document was labeled as positive sentiment, or why a word was classified as a noun---is less important than the accuracy of the prediction itself. 
The approaches we use and what we mean by `success' are thus guided by our research questions.

Domain experts and fellow researchers can provide feedback on questions and help with dynamically revising them. For example, they may say ``\textit{we already think we know that''}, ``\emph{that's too na\"ive}'', \emph{``that doesn't reflect social reality''} (negative); ``\textit{two major camps in the field would give different answers to that question}'' (neutral); ``\textit{we tried to look at that back in the 1960s, but we didn't have the technology}'' (positive); and ``\textit{that sounds like something that people who made that archive would love}'', ``\textit{that's a really fundamental  question}'' (very positive).

Sometimes we also hope to connect to multiple  disciplines. For example, while focusing on 
the humanistic concerns of an archive, we could also ask social questions such as ``\textit{is this archive more about collaborative processes, culture-building or norm creation?}'' or ``\textit{how well does this archive reflect the society in which it is embedded?}"
\cite{murdock2017exploration} used quantitative methods to tell a story about Darwin's intellectual development---an essential biographical question for a key figure in the history of science. At the same time, their methods connected Darwin's development to the changing landscape of Victorian scientific culture, allowing them to contrast Darwin's ``foraging'' in the scientific literature of his time to the ways in which that literature was itself produced. Finally, their methods provided a case study, and validation of technical approaches, for cognitive scientists who are interested in how people explore and exploit sources of knowledge. 

Questions about potential ``dual use'' may also arise. 
Returning to our introductory example,  \cite{chandrasekharan2017you} started with a deceptively simple question: if an internet platform eliminates forums for hate speech, does this impact hate speech in other forums? 
The research was motivated by the belief that a rising tide of online hate speech was (and is) making the internet increasingly unfriendly for disempowered groups, including minorities, women, and LBGTQ individuals. Yet the possibility of dual use  troubled the researchers from the onset. 
Could the methodology be adopted to target the speech of groups like Black Lives Matter? 
Could it be adopted by repressive governments to minimize online dissent? While these concerns remained, they  concluded that hypothetical dual use scenarios did not outweigh the tangible contribution this research could offer towards making the online environment more equal and just.

\section{Data}
\label{sec:data}
The next step involves deciding on the data sources, collecting and compiling the dataset, 
and inspecting its metadata.

\subsection{Data acquisition}
Many scholars in the humanities and the social sciences work with sources that are not available in digital form, and indeed may never be digitized. Others work with both analogue and digitized materials, and the  increasing digitization of archives has  opened opportunities to study these archives in new ways.
We can go to the canonical archive or open up something that nobody has studied before. For example, we might focus on major historical moments (French Revolution, post-Milosevic Serbia) or critical epochs (Britain entering the Victorian era, the transition from Latin to proto-Romance). Or, we could look for records of how people conducted science, wrote and consumed literature, and worked out their philosophies. 

\subsubsection{Born-digital data}
A growing number of researchers work with born-digital sources or data\footnote{See \cite{salganik2017bit} for an extensive discussion.}. 
Born-digital data, e.g., from social media, generally do not involve direct elicitation from participants and therefore enable unobtrusive measurements \citep{jaf2016tim,webb1966unobtrusive}. 
In contrast, methods like surveys sometimes elicit altered responses from participants, 
who  might adapt their responses to what they think is expected.
Moreover, born-digital data is often massive, enabling large-scale studies of language and behavior in a variety of social contexts.

Still, many scholars in the social sciences and humanities work with multiple data sources. The variety of sources typically used  means that more than one data collection method is often  required. 
For example,  a project examining coverage of a UK General Election, could draw data from  traditional media, web archives, Twitter and Facebook, campaign manifestos, etc. and might combine textual analysis of these materials with surveys, laboratory experiments, or field observations offline. 
In contrast, many computational studies based on born-digital data have focused on one specific source, such as Twitter.

The use of born-digital data raises  ethical  concerns.
Although early studies often treated privacy as a binary construct, many now acknowledge its complexity \citep{doi:10.1080/1369118X.2012.678878}. Conversations on private matters can be posted online, visible for all, but social norms
regarding what should be considered public information may differ from
the data's explicit visibility settings. Often no informed consent has been obtained, raising concerns and challenges regarding publishing content and potentially harmful secondary uses \citep{doi:10.1177/0038038517708140,salganik2017bit}.

Recently, concerns about potential harms stemming from secondary uses have led a number of digital service providers to restrict access to born-digital data. Facebook and Twitter, for example, have reduced or eliminated public access to their application programming interfaces (APIs) and expressed hesitation about allowing academic researchers to use data from their platforms to examine certain sensitive or controversial topics. Despite the seeming abundance of born-digital data, we therefore cannot take its availability for granted.

\subsubsection{Data quality}
Working with data that someone else has acquired 
presents additional problems related to provenance and contextualisation. It may not always be possible to determine the criteria applied during the creation process. For example, why were certain newspapers digitized but not others, and what does this 
say about the collection? Similar questions arise with the use of born-digital data. For instance, when using the Internet Archive’s Wayback Machine to gather data from archived web pages, we need to consider what pages were captured, which are likely missing, and why.

We must often repurpose born-digital data (e.g., Twitter was not designed to measure public opinion), but data biases may lead to spurious 
results and limit justification for generalization.\footnote{\cite{olteanu2016social} give an overview of biases in social data.} 
In particular, data collected via black box APIs designed for commercial, not research, purposes are likely to introduce biases into the inferences we draw, and the closed nature of these APIs means we rarely know what biases
are introduced, let alone how severely they might impact our research \citep{tromble2017we}. 
These, however, are not new problems. Historians, for example, have always understood that their sources were produced within particular contexts and for particular purposes, which are not always apparent to us.

Non-representative data can still be useful for making comparisons within a sample. 
In the introductory example  on hate speech \citep{chandrasekharan2017you}, the Reddit forums do not present a comprehensive or balanced picture of hate speech: the writing is almost exclusively in English, the targets of hate speech are mainly restricted (e.g., to black people, or women), and the population of writers is shaped by Reddit's demographics, which skew towards young white men. These biases limit the generalizability  of the findings, which cannot be extrapolated to other languages, other types of hate speech, and other demographic groups. However, because the findings are based on measurements on the same sort of hate speech and the same population of writers, as long as the collected data are representative of this specific population,  these biases do not pose an intractable validity problem if claims are properly restricted.

The size of many newly available datasets is one of their most appealing characteristics. Bigger datasets often make statistics more robust.  The size needed for a computational text analysis depends on the research goal: When it involves studying rare events, bigger datasets are needed. However,  larger is not always better. Some very large archives are  ``secretly'' collections of multiple and distinct processes that no in-field scholar would consider related.  For example, Google Books is frequently used to study cultural patterns, but the over-representation of scientific articles in Google books can be problematic \citep{10.1371/journal.pone.0137041}.  Even very large born-digital datasets usually cover limited timespans compared to, e.g., the Gutenberg archive of British novels.

This stage of the research also raises important questions about fairness. Are marginalized groups, for example, represented in the tweets we have collected? If not, what types of biases might result from analyses relying on those tweets?

Local experts and ``informants'' can help navigate the data. They can help understand the role an archive plays in the time and place.  They might tell us: Is this the central archive, or a peripheral one? What makes it unusual? 
Or they might tell us how certain underrepresented communities use a social media platform and advise us on strategies for ensuring our data collection includes their perspectives.

However, when it is practically infeasible to navigate the data in this way---for instance, when we cannot determine what is missing from Twitter's Streaming API or what webpages are left out of the Internet Archive---we should be open about the limitations of our analyses, acknowledging the flaws in our data and drawing cautious and reasonable conclusions from them. In all cases, we should report the choices we have made when creating or re-using any dataset.

\subsection{Compiling data}
After identifying the data source(s), the next step is compiling the data. 
This step is fundamental: if the sources cannot support a convincing result, no result will be convincing. 
In many cases, this involves defining a ``core" set of documents and a ``comparison" set.  We often have a specific set of documents in mind: an author's work, a particular journal, a time period. But if we want to say that this ``core" set has some distinctive property, we need a ``comparison" set. Expanding the collection beyond the documents that we would immediately think of has the beneficial effect of increasing our sample size. Having more sources increases the chance that we will notice something consistent across many individually varying contexts.

Comparing sets of documents can sometimes support  causal inference, presented as a contrast between a treatment group and a control. In \cite{chandrasekharan2017you}, the treatment consisted of the text written in the two forums that were eventually closed by Reddit. However, identifying a control group required a considerable amount of time and effort. Reddit is a diverse platform, with a wide variety of interactional and linguistic styles; it would be pointless to compare hate speech forums against forums dedicated to, say, pictures of wrecked bicycles.\footnote{https://www.reddit.com/r/bustedcarbon/} Chandrasekharan et al. used a matching design, populating the control group with forums that were as similar as possible to the treatment group, but were not banned from Reddit. 
The goal is to estimate the counterfactual scenario: in this case, what would have happened had the site not taken action against these specific forums?
An ideal control would make it possible to distinguish the effect of the treatment --- closing the forums --- from other idiosyncratic properties of texts that were treated.

We also look for categories of documents that might not be useful. We might remove documents that are meta-discourse, like introductions and notes, or documents that are in a language that is not the primary language of the collection, or duplicates when we are working with archived web pages.
However, we need to carefully consider the potential consequences of information we remove. Does its removal alter the data, or the interpretation of the data, we are analyzing? Are we losing anything that might be valuable at a later stage?

\subsection{Labels and metadata}
 Sometimes all we have is documents, but often we want to look at documents in the context of some additional information, or metadata. This additional information could tell us about the creation of documents (date, author, forum), or about the reception of documents (flagged as hate speech, helpful review).
  Information about text segments can be extremely valuable, but it is also prone to errors, inconsistencies, bias, and missing information. 
  Examining metadata is a good way to check a collection's balance and representativeness. Are sources disproportionately of one form? Is the collection missing a specific time window? This type of curation can be extremely time consuming as it may require expert labeling, but it often leads to the most compelling results. 
 Sometimes metadata are also used as target labels to develop machine learning models. 
 But using them as a ``ground truth''  requires caution. Labels sometimes mean something different than we expect. For example, a down vote for a social media post could indicate that the content is offensive, or  that the voter simply disagreed with the expressed view.

\section{Conceptualization}

\label{sec:conceptualization}
A core step in many analyses is translating social and cultural concepts (such as hate speech, rumor, or conversion) into measurable quantities. Before we can develop measurements for these concepts (the operationalization step, or the ``implementation'' step as denoted by \cite{piper2017think}), we need to define them. In the conceptualization phase we often start with questions such as: who are the domain experts, and how have they approached the topic? We are looking for a definition of the concept that is flexible enough to apply on our  dataset, yet formal enough for computational research. For example, our introductory study on hate speech \citep{chandrasekharan2017you} used  a statement on hate speech produced by  the European Union Court of Human Rights. The goal was not to implement this definition directly in software but to use it as a reference point to anchor  subsequent analyses.

If we want to move beyond the use of ad hoc definitions, 
it can be useful to distinguish between what political scientists  Adcock and Collier call the ``background concept'' and the ``systematized concept'' \citep{adcock2001measurement}.
The background concept comprises the full and diverse set of meanings that might be associated with a particular term. This involves  delving into theoretical, conceptual, and empirical studies to assess how a concept has been defined by  other scholars and, most importantly, to determine which definition is most appropriate for the particular research question and the theoretical framework in which it is situated. That definition, in turn, represents the systematized concept: the formulation that is adopted for the study. 

It is important to consider that for social and cultural concepts there is no absolute ground truth. There are often multiple valid definitions for a concept (the ``background'' concept in the terms of Adcock and Collier), and definitions might be contested over time. This may be uncomfortable for computer scientists, whose primary measure of success is often based on comparing a model's output  against ``ground truth'' or a ``gold standard'', e.g.,  by comparing a sentiment classifier's output against manual annotations.
However, the notion of ground truth is uncommon  in the humanities and the social sciences and it is often taken too far in machine learning.
\citet[p. 1]{kirschenbaum2007remaking} notes that in literary criticism and the digital humanities more broadly \emph{``interpretation, ambiguity, and argumentation are prized far
above ground truth and definitive conclusions"}. \citet[p. 2]{W13-1401} draw attention to the different attitudes of literary scholars and computational linguists towards ambiguity, stating that ``\textit{In Computational Linguistics [..] ambiguity
is almost uniformly treated as a problem to be
solved; the focus is on disambiguation, with the assumption
that one true, correct interpretation exists.}"
The latter is probably true for tasks such as spam filtering, but in the social sciences and the humanities many relevant concepts are fundamentally unobservable, such as latent traits of political actors \citep{lowe2013validating} or cultural fit in organizations \citep{doi:10.1287/mnsc.2016.2671}, leading to validation challenges. Moreover, when the  ground truth comes from people, it may be influenced by ideological priors, priming, simple differences of opinion or perspective, and many other factors \citep{doi:10.1177/2053951715602908}. We return to this issue in our discussions on validation and analysis.

\section{Operationalization}
\label{sec:operationalization}

In this phase we develop measures (or, ``operationalizations'', or ``indicators'') for the concepts of interest, a process called ``operationalization''. Regardless of whether we are working with computers,  the output produced coincides with Adcock and Collier's ``scores''---the concrete translation and output of the systematized concept into numbers or labels \citep{adcock2001measurement}.
Choices  made during this phase are  always tied to the question ``Are we measuring what we intend to measure?'' Does our operationalization match our conceptual definition? To ensure validity we must recognize gaps between what is important and what is easy to measure.
We first discuss  modeling considerations. Next, we describe several frequently used computational approaches and  their limitations and strengths.

\subsection{Modeling considerations}

\paragraph*{Variable types}

The variables (both predictors and outcomes)  are rarely simply  binary or categorical. For example, a study on language use and age could focus on chronological age (instead of, e.g., social age \citep{Eckert1997}). However, even then, age can be modeled in different ways. 
Discretization\footnote{Reducing a continuous variable to a discrete variable.} can make the modeling easier and various NLP studies  have modeled age as a categorical variable \citep{survey2015}. 
But any discretization raises questions: How many categories?  Where to place the boundaries?
 Fine distinctions might not always be meaningful for the analysis we are interested in, but categories that are too broad can threaten validity. 
 Other interesting variables include time, space, and even the social network position of the author.
 It is often preferable to keep the variable in its most precise form. 
 For example, \cite{nguyen2017kernel} perform exploration in the context of hypothesis testing by using latitude and longitude coordinates --- the original metadata attached to geotagged social media such as tweets --- rather than aggregating into administrative units such as counties or cities.
This is necessary when such administrative units are unlikely to be related to the target concept, as is the case in their analysis of dialect differences.
Focusing on precise geographical coordinates also makes it possible to recognize fine-grained effects, such as language variation across the geography of a city.

 \paragraph*{Categorization scheme} 
 Using a particular classification scheme means deciding which variations are visible, and which ones are hidden \citep{bowker1999sorting}. We are looking for a categorization scheme for which it is feasible to collect a large enough labeled document collection (e.g., to train supervised models), but which is also fine-grained enough for our purposes. Classification schemes rarely exhibit the ideal properties, i.e., that they are consistent, their categories are mutually exclusive, and that the system is complete  \citep{bowker1999sorting}.   
Borderline cases are  challenging, especially with social and cultural concepts, where the boundaries are often not clear-cut.  The choice of scheme can also have ethical implications \citep{bowker1999sorting}. For example, gender is usually represented as a binary variable  in NLP  and computational models tend to learn gender-stereotypical patterns. The operationalization of gender in NLP has been challenged only recently \citep{bamman2014gender,koolen2017these,nguyen142014gender}.

\paragraph*{Supervised vs. unsupervised} 
Supervised and unsupervised learning are the most common approaches to learning from data. With supervised learning, a model learns from \emph{labeled} data (e.g., social media messages labeled by sentiment) to infer (or predict) these labels from unlabeled texts.
 In contrast, unsupervised learning uses \emph{unlabeled} data. Supervised approaches are especially suitable when we have a clear definition of the concept of interest and when labels are available (either annotated or native to the data). 
Unsupervised  approaches, such as topic models, are  especially useful for exploration. In this setting, conceptualization and operationalization may occur simultaneously, with theory emerging from the data \citep{doi:10.1002/asi.23786}. 
Unsupervised approaches are  also used when there is a clear way of measuring a concept, often based on strong assumptions. For example, \cite{murdock2017exploration} measure ``surprise'' in an analysis of Darwin's reading decisions based on the divergence between two probability distributions.

\paragraph*{Units of interest}
From an analysis perspective, the unit of text that we are labeling (or annotating, or coding), either automatic or manual, can sometimes be different than one's final unit of analysis.  For example, if in a study on media frames in news stories, the theoretical framework and research question point toward frames at the story level (e.g., what is the overall causal analysis of the news article?), the story must be the unit of analysis. Yet it is often difficult to validly and reliably code a single frame at the story level. Multiple perspectives are likely to sit side-by-side in a story. Thus, an article on income inequality might point to multiple causes, such as globalization, education, and tax policies. Coding at the sentence level would detect each of these causal explanations individually, but this information would need to be somehow aggregated to determine the overall story-level frame. Sometimes scholars solve this problem by  only examining headlines and lead paragraphs, arguing that based on journalistic convention, the most important information can be found at the beginning of a story. However, this leads to a return to a shorter, less nuanced analysis.

From a computational perspective, the unit of text can also make  a huge difference, especially when we are using bag-of-words models, where word order within a unit does not matter. 
Small segments, like tweets, sometimes do not have enough information to make their semantic context clear. In contrast,  larger segments, like novels, have too much variation, making it difficult to train focused models. Finding a good segmentation sometimes means combining short documents and subdividing long documents. The word ``document" can therefore be misleading. But it is so ingrained in the common NLP lexicon that we use it anyway in this article.

\paragraph*{Interpretability}
For insight-driven text analysis, it is often critical that high-level patterns can be communicated. Furthermore, interpretable models make it easier to find spurious features, to do error analysis, and to support interpretation of results. Some approaches  are  effective  for prediction, but harder to interpret.  The value we place on interpretability can therefore influence the approach we choose. There is an increasing interest in developing interpretable or transparent models in the NLP and machine learning communities.\footnote{e.g. the  Fairness, Accountability, and Transparency (FAT$^*$) conference.}

\subsection{Annotation}
Many studies involve human coders. Sometimes the goal is to fully code the data, but in a computational analysis we often use the labels (or annotations)  to train machine learning models to automatically recognize them, and to identify language patterns that are associated with these labels. 
For example, for a project analyzing rumors online \citep{zubiaga2016analysing}, conversation threads were annotated along different dimensions, including rumor versus non-rumor and  stance towards a rumor.

The collection of annotation choices make up an annotation scheme (or ``codebook''). 
Existing schemes and annotations can be useful as starting points. 
Usually settling on an annotation scheme requires several iterations, in which the guidelines are updated and annotation examples are added. 
For example, a political scientist could use a mixed deductive-inductive strategy for developing a codebook. She starts by laying out a set of theory-driven deductive coding rules, which means that the broad principles of the coding rules are laid out without examining examples first. These are then tested (and possibly adjusted) based on a  sample of the data. 
 In line with Adcock and Collier's notion of ``content validity'' \citep{adcock2001measurement}, the goal is to assess whether the codebook adequately captures the systematized concept. By looking at the data themselves, she gains  a better sense of whether some things have been left out of the coding rules and whether anything is superfluous, misleading, or confusing. Adjustments are made and the process is  repeated, often with another researcher involved.

The final annotations can be collected using a crowdsourcing platform, a smaller number of highly-trained annotators, or a group of experts. Which type of annotator to use should be informed by the complexity and specificity of the concept. For more complex concepts, highly-trained or expert annotators tend to produce more reliable results. However, complex concepts can sometimes be broken down into micro-tasks that can be performed independently in parallel by crowdsourced annotators. Concepts from highly specialized domains may require expert annotators. In all cases, however, some training will be required, and the training phase should involve continual checks of inter-annotator agreement (i.e. intercoder reliability) or checks against a gold standard (e.g. quizzes in crowdsourcing platforms).

We also need to decide how inter-annotator agreement will be measured and what an acceptable level of agreement would be.  Krippendorff's alpha is frequently used in the social sciences, but the right measure depends on the type of data and task.
For manual coding, we can continually check  inter-annotator agreement and begin introducing checks of  \textit{intra}-annotator agreement, too. 
For most communication scholars using only manual content analysis, an acceptable rate of agreement is achieved when Krippendorf's alpha reaches 0.80 or above. 
When human-coded data are used to  validate machine learning algorithms, the reliability of the human-coded data is even more important. 
Disagreement between annotators can  signal   weaknesses of the annotation scheme, or highlight the inherent ambiguity in what we are trying to measure. Disagreement itself can be meaningful and can be integrated in subsequent analyses   \citep{aroyo2013crowd,Demeester2016}.

\subsection{Data pre-processing}
Preparing the data can be a complex and time-consuming process, often involving working with partially or wholly unstructured data. 
The pre-processing steps have a big impact on the operationalizations, subsequent analyses and reproducibility efforts \citep{fokkens-EtAl:2013:ACL2013}, and they are usually tightly linked to what we intend to measure. Unfortunately, these steps tend to be underreported, but documenting the pre-processing choices made is essential and is analogous to recording the decisions taken during the production of a scholarly edition or protocols in biomedical research.
Data may also vary enormously in quality, depending on how it has been generated. Many historians, for example, work with text produced from an analogue original using Optical Character Recognition (OCR). Often, there will be limited information available regarding the accuracy of the OCR, and the degree of accuracy may even vary within a single corpus (e.g. where digitized text has been produced over a period of years, and the software has gradually improved). The first step, then, is to try to correct for common OCR errors. These will vary depending on the type of text, the date at which the `original' was produced, and the nature of the font and typesetting.

One  step that almost everyone takes  is to tokenize the original character sequence into  the words and word-like units. Tokenization is a more subtle and more powerful process than people expect.  It is often done using regular expressions or scripts that have been circulating within the NLP community. Tokenization heuristics, however, can be  badly confused by emoticons, creative orthography (e.g., U\$A, sh!t), and missing whitespace. 
Multi-word terms are also challenging. Treating them as a single unit can dramatically alter the patterns in text. Many words that are individually ambiguous have clear, unmistakable meanings as terms, like ``black hole" or ``European Union". However, deciding what constitutes a multi-word term is a difficult problem. 
In writing systems like Chinese, tokenization is a research problem in its own right. 

Beyond tokenization, common  steps include lowercasing,  removing punctuation, stemming (removing suffixes), lemmatization (converting inflections to a base lemma), and normalization, which has never been clearly defined\footnote{\citep{sproat2001} make a good first step, but this work precedes the wave of research on social media, and does not include many things that are considered normalization today.}, but often includes grouping abbreviations like ``U.S.A." and ``USA", ordinals like ``1st" and ``first", and variant spellings like ``noooooo". The main goal of these  steps is to improve the ratio of tokens (individual occurrences) to types (the distinct things in a corpus). 
Each step requires making additional assumptions about which distinctions are relevant: is ``apple'' different from ``Apple''? Is ``burnt'' different from ``burned''? Is ``cool" different from ``coooool"?
Sometimes these steps can actively hide useful patterns, like social meaning \citep{eisenstein:2013:NAACL-HLT}. Some of us  therefore try do as little modification as possible.

From a multilingual perspective, English and Chinese have an unusually simple inflectional system, and so it is statistically reasonable to treat each inflection as a unique word type. Romance languages have considerably more inflections than English; many indigenous North American languages have still more. 
For these languages, unseen data is far more likely to include previously-unseen inflections, and therefore, dealing with inflections is more important. On the other hand, the resources for handling inflections vary greatly by language, with European languages dominating the attention of the computational linguistics community thus far.

We sometimes also remove words that are not relevant to our goals, for example by calculating vocabulary frequencies. We construct a ``stoplist'' of words that we are not interested in. If we are looking for semantic themes we might remove function words like determiners and prepositions. If we are looking for author-specific styles, we might remove all words except function words. Some words are generally meaningful but too frequent to be useful within a specific collection. 
We sometimes also remove very infrequent words. Their occurrences are too low for robust patterns and removing them  helps reducing the vocabulary size.

The choice of processing steps can be guided by theory or knowledge about the domain as well as experimental investigation. When we have labels, predictive accuracy of a model is a way to assess the effect of the processing steps. 
 In unsupervised settings, it is more challenging to understand the effects of different steps. Inferences drawn from unsupervised settings can be sensitive to  pre-processing choices  \citep{denny-spirling-2018}. 
 Stemming has been found to provide little measurable benefits for topic modeling and can sometimes even be harmful \citep{TACL868}. 
 All in all, this again highlights the need to document these steps.

Finally, we can also mark up the data, e.g.,  by identifying entities  (people, places, organizations, etc.) or parts of speech. 
Although many NLP tools are available for such tasks, they are often challenged by linguistic variation, such as orthographic variation in historical texts  \citep{doi:10.2200/S00436ED1V01Y201207HLT017} and social media \citep{eisenstein:2013:NAACL-HLT}. Moreover, the performance of NLP tools often drops when applying them outside the training domain, such as applying tools developed on newswire texts to texts written by younger authors \citep{hovyage2015}.  
Problems (e.g., disambiguation in named entity recognition) are sometimes resolved using considerable manual intervention. This combination of the automated and the manual, however, becomes more difficult as the scale of the data increases, and the `certainty' brought by the latter may have to be abandoned.

\subsection{Dictionary-based approaches}
Dictionaries are frequently used to code texts in  content analyses \citep{neuendorf2016content}.
Dictionaries consist of one or more categories (i.e. word lists). Sometimes the output is simply the number of category occurrences (e.g., positive sentiment), thus weighting words within a category equally. In some other cases, words are assigned continuous scores.
The  high transparency of dictionaries makes them sometimes more suitable than  supervised machine learning models.  However, dictionaries should only be used if the scores assigned to words 
match how the words are used in the data  (see \cite{grimmer-stewart-2013} for a detailed discussion on limitations). 
There are many off-the-shelf dictionaries available (e.g., LIWC \citep{liwc2010}).
 These are often well-validated, but applying them on a new domain may not be appropriate without additional validation. Corpus- or domain-specific dictionaries can overcome limitations of general-purpose dictionaries. 

The dictionaries are often manually compiled, but increasingly they are constructed semi-automatically (e.g., \cite{fast2016empath}). 
When we semi-automatically create a word list, we use automation to identify an initial word list, and human insight to filter it. By automatically generating the initial words lists, words can be identified that human annotators might have difficulty intuiting. By manually filtering the lists, we use our theoretical understanding of the target concept to remove spurious features.

In the introduction study, 
SAGE \citep{eisenstein2011sparse} was used to obtain a list of words that distinguished the text in the treatment group (subreddits that were closed by Reddit) from text in the control group (similar subreddits that were not closed). 
The researchers then returned to the hate speech definition provided by the European Court of Human Rights, and manually filtered the top SAGE words based on this definition. Not all identified words fitted the definition. The others included: the names of the subreddits themselves, names of related subreddits, community-specific jargon that was not directly related to hate speech, and terms such as \textit{IQ} and \textit{welfare}, which were frequently used in discourses of hate speech, but had significant other uses. The word lists provided the measurement instrument for their main result, which is that the use of hate speech throughout Reddit declined after the two treatment subreddits were closed. 

\subsection{Supervised models}
Supervised learning is frequently used to scale up analyses. For example,  \cite{nguyenemnlp2015}  wanted to analyze the motivations of  Movember campaign participants.
 By developing a classifier based on a small set of annotations, they were able to expand the analysis to over 90k participants. 
 
The choice of supervised learning model  is often guided by the task definition and the label types.
 For example, to identify stance towards rumors  based on sequential annotations, an algorithm for learning from sequential \citep{zubiaga-EtAl:2016:COLING} or time series data \citep{lukasik-EtAl:2016:P16-2} could be used.  
The features (sometimes called variables or predictors) are used by the model to make the predictions. 
They may vary from content-based features such as single words,  sequences of words, or information about their syntactic structure,  to meta-information such as user or network information. Deciding on the  features  requires experimentation and expert insight and is often called feature engineering. 
For insight-driven analysis, we are often  interested in \emph{why} a prediction has been made and features that can be interpreted by humans may be preferred. Recent neural network approaches often use simple features as input (such as word embeddings or character sequences), which requires  less feature engineering but make  interpretation more difficult. 

Supervised models are powerful, but they can latch on to spurious features of the dataset. This is particularly true for datasets that are not well-balanced, and for annotations that are noisy. In our introductory example on hate speech in Reddit \citep{chandrasekharan2017you}, the annotations are automatically derived from the forum in which each post appears, and indeed, many of the posts in the forums (subreddits) that were banned by Reddit would 
be perceived by many as hate speech. But even in banned subreddits, not all of the content is hate speech (e.g., some of the top features were self-referential  like the name of the subreddit) but a classifier would learn a high weight for these features.

Even when expert annotations are available on the level of individual posts, spurious features may remain. \cite{waseem-hovy:2016:N16-2} produced expert annotations of hate speech on Twitter. They found that one of the strongest features for sexism is the name of an Australian TV show, because people like to post sexist comments about the contestants. If we are trying to make claims about what inhibits or encourages hate speech, we would not want those claims to be tied to the TV show's popularity. 
Such problems are inevitable when datasets are not well-balanced over time, across genres, topics, etc. Especially with social media data, we lack a clear and objective definition of `balance' at this time.
 
The risk of supervised models latching on to spurious features reinforces the need for interpretability. Although the development of supervised models is usually performance driven,  placing more emphasis on interpretability could  increase the adoption of these models in insight-driven analyses.  
One way would be to only use  models that are already somewhat   interpretable, for example models that use a small number of  human-interpretable features. Rather than imposing such restrictions, 
there is also work on generating post-hoc  explanations for individual predictions  (e.g., \cite{Ribeiro:2016:WIT:2939672.2939778}), even when the underlying model itself is very complex.

\subsection{Topic modeling}
Topic models (e.g., LDA \citep{blei2003latent}) are  usually unsupervised and therefore less biased towards human-defined categories. They are  especially suited for insight-driven analysis,  because they are constrained in ways that make their output interpretable. Although there is no guarantee that a  ``topic'' will correspond to a recognizable theme or event or discourse, they often do so in ways that other methods do not. Their easy applicability without supervision and ready interpretability make topic models good for exploration.
Topic models are less successful for many performance-driven applications. Raw word features are almost always better than topics for search and document classification. LSTMs and other neural network models are better as language models. Continuous word embeddings have more expressive power to represent fine-grained semantic similarities between words.

A topic model provides a different perspective on a collection. It creates a set of probability distributions over the vocabulary of the collection, which, when combined together in different proportions, best match the content of the collection. We can sort the words in each of these distributions in descending order by probability, take some arbitrary number of most-probable words, and get a sense of what (if anything) the topic is ``about''. Each of the text segments also has its own distribution over the topics, and we can sort these segments by their probability within a given topic to get a sense of how that topic is used.

One of the most common questions about topic models is how many topics to use, usually with the implicit assumption that there is a ``right'' number that is inherent in the collection. We prefer to think of this parameter as more like the scale of a map or the magnification of a microscope. The ``right'' number is determined by the needs of the user, not by the collection. If the analyst is looking for a broad overview, a relatively small number of topics may be best. If the analyst is looking for fine-grained phenomena, a larger number is better.

After fitting the model, it may be necessary to circle back to an earlier phase. Topic models find consistent patterns. When authors repeatedly use a particular theme or discourse, that repetition creates a consistent pattern. But other factors can also create similar patterns, which look as good to the algorithm. We might notice a topic that has highest probability on French stopwords, indicating that we need to do a better job of filtering by language. We might notice a topic of word fragments, such as ``\textit{ing}'', ``\textit{tion}'', ``\textit{inter}'', indicating that we are not handling end-of-line hyphenation correctly. We may need to add to our stoplist or change how we curate multi-word terms.

\subsection{Validation}
The output of our measurement procedures  (in the social sciences often called the ``scores'') must now be assessed in terms of their reliability and validity with regard to the (systemized) concept. Reliability aims to capture repeatability, i.e. the extent to which a given tool provides consistent results. 

Validity assesses the extent to which a given measurement tool measures what it is supposed to measure. 
In NLP and machine learning, most models are primarily evaluated  by  comparing the  machine-generated labels  against an annotated sample. This approach presumes that the human output is the ``gold standard" against which performance should be tested. In contrast, 
when the reliability is measured based on the output of different annotators, no coder is taken as the standard and  the likelihood of coders reaching agreement by chance (rather than because they are ``correct") is factored into the resulting statistic.  Comparing against a ``gold standard''  suggests that the threshold for human inter- and intra-coder reliability should be particularly high.

Accuracy, as well as other measures such as precision, recall and F-score, are sometimes presented as a measure of validity, but if we do not have a genuinely objective determination of what something is supposed measure---as is often the case in text analysis---then accuracy is perhaps a better indication of reliability than of validity. In that case, validity needs to be assessed based on other  techniques like those we discuss later in this section.  It is also worth asking  what level of accuracy is sufficient for our analysis and to what extent there may be an upper bound, especially when the labels are native to the data or when the notion of a ``gold standard'' is not appropriate. 

For some in the humanities, validation takes the form of close reading, not designed to confirm whether the model output is correct, but to present what  \citet[pp. 67-68]{piper2015novel} refers to as a form of ``\textit{further discovery in two directions}''. Model outputs tell us something about the texts, while a close reading of the texts alongside those outputs tells us something about the models that can be used for more effective model building. Applying this circular, iterative process to 450 18th-century novels written in three languages, Piper was able to uncover a new form of ``\textit{conversional novel}'' that was not previously captured in ``\textit{literary history's received critical categories}'' \citep[p. 92]{piper2015novel}.

Along similar lines,  we can subject both the machine-generated output and the human annotations  to another round of content validation. That is, take a stratified random sample, selecting observations from the full range of scores, and ask: Do these make sense in light of the systematized concept? If not, what seems to be missing? Or is something extraneous being captured? This is primarily a qualitative process that requires returning to theory and interrogating the systematized concept, indicators, and scores together.
This type of validation is rarely done in NLP, but it is especially important when it is difficult to assess what drives a given machine learning model.
If there is a mismatch between the scores and systematized concept at this stage, the codebook may need to be adjusted, human coders retrained, more training data prepared, algorithms adjusted, or in some instances, even a new analytical method adopted.

Other types of validation are also possible, such as comparing with  
other approaches that aim to capture the same concept, or comparing the output with external measures (e.g., public opinion polls, the occurrence of future events).
We can also go beyond only evaluating the labels (or point estimates). \cite{lowe2013validating} 
used human judgments to not only assess the positional estimates from a scaling method  of latent political traits but  also to assess uncertainty intervals. 
Using different types of validation can increase our
confidence in the approach, 
especially when there is no clear notion of ground truth. 

Besides focusing on rather abstract evaluation measures, we could also assess  the models in task-based settings  using human experts. Furthermore, for insight-driven analyses, it can be more useful to focus on improving explanatory power  than making small improvements in predictive performance.

\section{Analysis}
\label{sec:analysis}
In this phase, we use our models to explore or answer our research questions.  
For example, given a topic model we can look at the connection between 
 topics and metadata elements. Tags such as ``hate speech" or metadata information imply a certain way of organizing the collection. Computational models provide another organization, which may differ in ways that provide more insight into how these categories manifest themselves, or fail to do so. 

Moreover, when using a supervised approach,  the ``errors'', i.e.  disagreement 
between the system output and human-provided labels,  can point towards interesting cases for closer analysis and help us reflect on our conceptualizations. In  the words of \cite{long2016literary}, they can be ``\emph{opportunities for interpretation}''. 
Other types of ``failures'' can be insightful as well. Sometimes there is a ``\textit{dog that didn't bark}'' \citep{holmes1892}--i.e., something that everyone thinks we should have found, but we did not. Or, sometimes the failures are  telling us about the existence of something in the data that nobody noticed, or thought important, until then (e.g., the large number of travel journals in Darwin's reading lists). 

 Computational text analysis is not a replacement for but rather an addition to the approaches one can take to analyze social and cultural phenomena using textual data. By moving back and forth between large-scale computational analyses and small-scale qualitative analyses, we can combine their strengths so that we can identify large-scale and long-term trends, but also tell individual stories. 
For example, the Reddit study on hate speech \citep{chandrasekharan2017you} raised various  follow-up questions: Can we distinguish  hate speech from people talking about hate speech? Did  people  find new ways to express hate speech? If so, did the total amount of online hate speech decrease after all?
As possible next steps, a qualitative discourse analyst might examine a smaller corpus to investigate whether commenters were indeed expressing hate speech in new ways; a specialist in interview methodologies might reach out to commenters to better understand the role of online hate speech in their lives. Computational text analysis represents a step towards better understanding social and cultural phenomena, and it is in many cases better suited towards opening questions rather than closing them.

\section{Conclusion}
\label{sec:conclusion}
Insight-driven computational  analysis of text is becoming increasingly common.
It not only helps us see more broadly, it  helps us see 
subtle patterns more clearly and allows us to explore radical new questions about culture and society. 
In this article we have consolidated our experiences,
as scholars from very different disciplines, in analyzing text as social and cultural data and described how the research process often unfolds.
Each of the steps in the process is time-consuming and labor-intensive. Each presents challenges. And especially when working across disciplines, the research often involves a fair amount of  discussion---even negotiation---about what means of operationalization and approaches to analysis  are appropriate and feasible. 
And yet, with a bit of perseverance and mutual understanding, conceptually sound and meaningful work results so that we
can truly make use of  the exciting opportunities  rich textual data offers.

\section{Acknowledgements}
This work was supported by The Alan Turing
Institute under the EPSRC grant EP/N510129/1.
Dong Nguyen is supported with an Alan Turing Institute Fellowship (TU/A/000006). Maria Liakata is a Turing fellow at 40\%. We would also like to thank the participants of the
``Bridging disciplines in analysing text as social and cultural data'' workshop held at the Turing Institute (2017) for insightful discussions.  The workshop was funded by a Turing Institute seed funding award to Nguyen and Liakata.

\bibliographystyle{plainnat}
\bibliography{bib_dong}

\begin{thebibliography}{50}
\providecommand{\natexlab}[1]{#1}
\providecommand{\url}[1]{\texttt{#1}}
\expandafter\ifx\csname urlstyle\endcsname\relax
  \providecommand{\doi}[1]{doi: #1}\else
  \providecommand{\doi}{doi: \begingroup \urlstyle{rm}\Url}\fi

\bibitem[Adcock and Collier(2001)]{adcock2001measurement}
Robert Adcock and David Collier.
\newblock Measurement validity: A shared standard for qualitative and
  quantitative research.
\newblock \emph{American Political Science Review}, 95\penalty0 (3):\penalty0
  529--546, 2001.

\bibitem[Aroyo and Welty(2013)]{aroyo2013crowd}
Lora Aroyo and Chris Welty.
\newblock Crowd truth: Harnessing disagreement in crowdsourcing a relation
  extraction gold standard.
\newblock In \emph{Proceedings of WebSci'13}, 2013.

\bibitem[Bamman et~al.(2014)Bamman, Eisenstein, and
  Schnoebelen]{bamman2014gender}
David Bamman, Jacob Eisenstein, and Tyler Schnoebelen.
\newblock Gender identity and lexical variation in social media.
\newblock \emph{Journal of Sociolinguistics}, 18\penalty0 (2):\penalty0
  135--160, 2014.

\bibitem[Baumer et~al.(2017)Baumer, Mimno, Guha, Quan, and
  Gay]{doi:10.1002/asi.23786}
Eric P.~S. Baumer, David Mimno, Shion Guha, Emily Quan, and Geri~K. Gay.
\newblock Comparing grounded theory and topic modeling: Extreme divergence or
  unlikely convergence?
\newblock \emph{Journal of the Association for Information Science and
  Technology}, 68\penalty0 (6):\penalty0 1397--1410, 2017.

\bibitem[Blei et~al.(2003)Blei, Ng, and Jordan]{blei2003latent}
David~M. Blei, Andrew~Y. Ng, and Michael~I. Jordan.
\newblock Latent {{Dirichlet}} allocation.
\newblock \emph{Journal of machine Learning research}, 3:\penalty0 993--1022,
  2003.

\bibitem[Bowker and Star(1999)]{bowker1999sorting}
Geoffrey~C. Bowker and Susan~Leigh Star.
\newblock \emph{Sorting Things Out: Classification and Its Consequences}.
\newblock MIT Press, 1999.

\bibitem[Chandrasekharan et~al.(2017)Chandrasekharan, Pavalanathan, Srinivasan,
  Glynn, Eisenstein, and Gilbert]{chandrasekharan2017you}
Eshwar Chandrasekharan, Umashanthi Pavalanathan, Anirudh Srinivasan, Adam
  Glynn, Jacob Eisenstein, and Eric Gilbert.
\newblock You can't stay here: The effectiveness of {Reddit's} 2015 ban through
  the lens of hate speech.
\newblock \emph{Proceedings of the {ACM} on Human-Computer Interaction},
  1\penalty0 (2), 2017.

\bibitem[danah boyd and Crawford(2012)]{doi:10.1080/1369118X.2012.678878}
danah boyd and Kate Crawford.
\newblock Critical questions for big data: Provocations for a cultural,
  technological, and scholarly phenomenon.
\newblock \emph{Information, Communication \& Society}, 15\penalty0
  (5):\penalty0 662--679, 2012.

\bibitem[Demeester et~al.(2016)Demeester, Aly, Hiemstra, Nguyen, and
  Develder]{Demeester2016}
Thomas Demeester, Robin Aly, Djoerd Hiemstra, Dong Nguyen, and Chris Develder.
\newblock Predicting relevance based on assessor disagreement: Analysis and
  practical applications for search evaluation.
\newblock \emph{Information Retrieval Journal}, 19\penalty0 (3):\penalty0
  284--312, 2016.

\bibitem[Denny and Spirling(2018)]{denny-spirling-2018}
Matthew~J. Denny and Arthur Spirling.
\newblock Text preprocessing for unsupervised learning: Why it matters, when it
  misleads, and what to do about it.
\newblock \emph{Political Analysis}, 26\penalty0 (2):\penalty0 168--189, 2018.

\bibitem[DiMaggio(2015)]{doi:10.1177/2053951715602908}
Paul DiMaggio.
\newblock Adapting computational text analysis to social science (and vice
  versa).
\newblock \emph{Big Data \& Society}, 2015.

\bibitem[Doyle(1892)]{holmes1892}
A.~Conan Doyle.
\newblock Adventures of {{Sherlock Homes}}: The adventure of {{Silver Blaze}}.
\newblock \emph{Strand Magazine}, Vol. IV, December 1892:\penalty0 291--306,
  1892.

\bibitem[Eckert(1997)]{Eckert1997}
Penelope Eckert.
\newblock Age as a sociolinguistic variable.
\newblock In Florian Coulmas, editor, \emph{The Handbook of Sociolinguistics},
  pages 151--167. Blackwell Publishers, 1997.

\bibitem[Eisenstein(2013)]{eisenstein:2013:NAACL-HLT}
Jacob Eisenstein.
\newblock What to do about bad language on the internet.
\newblock In \emph{Proceedings of the 2013 Conference of the North American
  Chapter of the Association for Computational Linguistics: Human Language
  Technologies}, pages 359--369, 2013.

\bibitem[Eisenstein et~al.(2011)Eisenstein, Ahmed, and
  Xing]{eisenstein2011sparse}
Jacob Eisenstein, Amr Ahmed, and Eric~P. Xing.
\newblock Sparse additive generative models of text.
\newblock In \emph{{Proceedings of the International Conference on Machine
  Learning (ICML)}}, pages 1041--1048, 2011.

\bibitem[Fast et~al.(2016)Fast, Chen, and Bernstein]{fast2016empath}
Ethan Fast, Binbin Chen, and Michael~S. Bernstein.
\newblock Empath: Understanding topic signals in large-scale text.
\newblock In \emph{Proceedings of the 2016 CHI Conference on Human Factors in
  Computing Systems}, pages 4647--4657, 2016.

\bibitem[Fokkens et~al.(2013)Fokkens, van Erp, Postma, Pedersen, Vossen, and
  Freire]{fokkens-EtAl:2013:ACL2013}
Antske Fokkens, Marieke van Erp, Marten Postma, Ted Pedersen, Piek Vossen, and
  Nuno Freire.
\newblock Offspring from reproduction problems: What replication failure
  teaches us.
\newblock In \emph{Proceedings of the 51st Annual Meeting of the Association
  for Computational Linguistics (Volume 1: Long Papers)}, pages 1691--1701,
  2013.

\bibitem[Grimmer and Stewart(2013)]{grimmer-stewart-2013}
Justin Grimmer and Brandon~M. Stewart.
\newblock Text as data: The promise and pitfalls of automatic content analysis
  methods for political texts.
\newblock \emph{Political Analysis}, 21\penalty0 (3):\penalty0 267--297, 2013.

\bibitem[Hammond et~al.(2013)Hammond, Brooke, and Hirst]{W13-1401}
Adam Hammond, Julian Brooke, and Graeme Hirst.
\newblock A tale of two cultures: Bringing literary analysis and computational
  linguistics together.
\newblock In \emph{Proceedings of the Workshop on Computational Linguistics for
  Literature}, pages 1--8, 2013.

\bibitem[Hovy and S{\o}gaard(2015)]{hovyage2015}
Dirk Hovy and Anders S{\o}gaard.
\newblock Tagging performance correlates with author age.
\newblock In \emph{Proceedings of the 53rd Annual Meeting of the Association
  for Computational Linguistics and the 7th International Joint Conference on
  Natural Language Processing (Volume 2: Short Papers)}, pages 483--488,
  Beijing, China, 2015.

\bibitem[Kirschenbaum(2007)]{kirschenbaum2007remaking}
Matthew~G. Kirschenbaum.
\newblock The remaking of reading: Data mining and the digital humanities.
\newblock In \emph{The National Science Foundation Symposium on Next Generation
  of Data Mining and Cyber-Enabled Discovery for Innovation, Baltimore, MD},
  2007.

\bibitem[Koolen and van Cranenburgh(2017)]{koolen2017these}
Corina Koolen and Andreas van Cranenburgh.
\newblock These are not the stereotypes you are looking for: Bias and fairness
  in authorial gender attribution.
\newblock In \emph{Proceedings of the First Workshop on Ethics in Natural
  Language Processing}, pages 12--22, 2017.

\bibitem[Long and So(2016)]{long2016literary}
Hoyt Long and Richard~Jean So.
\newblock Literary pattern recognition: Modernism between close reading and
  machine learning.
\newblock \emph{Critical Inquiry}, 42\penalty0 (2):\penalty0 235--267, 2016.

\bibitem[Lowe and Benoit(2013)]{lowe2013validating}
Will Lowe and Kenneth Benoit.
\newblock Validating estimates of latent traits from textual data using human
  judgment as a benchmark.
\newblock \emph{Political Analysis}, 21\penalty0 (3):\penalty0 298--313, 2013.

\bibitem[Lukasik et~al.(2016)Lukasik, Srijith, Vu, Bontcheva, Zubiaga, and
  Cohn]{lukasik-EtAl:2016:P16-2}
Michal Lukasik, P.~K. Srijith, Duy Vu, Kalina Bontcheva, Arkaitz Zubiaga, and
  Trevor Cohn.
\newblock Hawkes processes for continuous time sequence classification: an
  application to rumour stance classification in {Twitter}.
\newblock In \emph{Proceedings of the 54th Annual Meeting of the Association
  for Computational Linguistics (Volume 2: Short Papers)}, pages 393--398,
  Berlin, Germany, 2016.

\bibitem[Mosteller and Wallace(1963)]{doi:10.1080/01621459.1963.10500849}
Frederick Mosteller and David~L. Wallace.
\newblock Inference in an authorship problem.
\newblock \emph{Journal of the American Statistical Association}, 58\penalty0
  (302):\penalty0 275--309, 1963.

\bibitem[Murdock et~al.(2017)Murdock, Allen, and DeDeo]{murdock2017exploration}
Jaimie Murdock, Colin Allen, and Simon DeDeo.
\newblock Exploration and exploitation of {Victorian Science in Darwin's}
  reading notebooks.
\newblock \emph{Cognition}, 159:\penalty0 117--126, 2017.

\bibitem[Neuendorf(2002)]{neuendorf2016content}
Kimberly~A. Neuendorf.
\newblock \emph{The Content Analysis Guidebook}.
\newblock Sage, 2002.

\bibitem[Nguyen and Eisenstein(2017)]{nguyen2017kernel}
Dong Nguyen and Jacob Eisenstein.
\newblock A kernel independence test for geographical language variation.
\newblock \emph{Computational Linguistics}, 43\penalty0 (3):\penalty0 567--592,
  2017.

\bibitem[Nguyen et~al.(2014)Nguyen, Trieschnigg, Do{\u{g}}ru{\"o}z, Gravel,
  Theune, Meder, and de~Jong]{nguyen142014gender}
Dong Nguyen, Dolf Trieschnigg, A.~Seza Do{\u{g}}ru{\"o}z, Rilana Gravel,
  Mari{\"e}t Theune, Theo Meder, and Franciska de~Jong.
\newblock Why gender and age prediction from tweets is hard: Lessons from a
  crowdsourcing experiment.
\newblock In \emph{Proceedings of COLING 2014, the 25th International
  Conference on Computational Linguistics: Technical Papers}, pages 1950--1961,
  Dublin, Ireland, 2014.

\bibitem[Nguyen et~al.(2015)Nguyen, van~den Broek, Hauff, Hiemstra, and
  Ehrenhard]{nguyenemnlp2015}
Dong Nguyen, Thijs van~den Broek, Claudia Hauff, Djoerd Hiemstra, and Michel
  Ehrenhard.
\newblock \#{S}upportthecause: Identifying motivations to participate in online
  health campaigns.
\newblock In \emph{Proceedings of the 2015 Conference on Empirical Methods in
  Natural Language Processing}, pages 2570--2576, Lisbon, Portugal, 2015.

\bibitem[Nguyen et~al.(2016)Nguyen, Do{\u{g}}ru{\"o}z, Ros{\'e}, and
  de~Jong]{survey2015}
Dong Nguyen, A.~Seza Do{\u{g}}ru{\"o}z, Carolyn~P. Ros{\'e}, and Franciska
  de~Jong.
\newblock Computational sociolinguistics: A survey.
\newblock \emph{Computational Linguistics}, 42\penalty0 (3):\penalty0 537--593,
  2016.

\bibitem[Olteanu et~al.(2016)Olteanu, Castillo, Diaz, and
  Kiciman]{olteanu2016social}
Alexandra Olteanu, Carlos Castillo, Fernando Diaz, and Emre Kiciman.
\newblock Social data: Biases, methodological pitfalls, and ethical boundaries.
\newblock \emph{Available at SSRN: https://ssrn.com/abstract=2886526}, 2016.

\bibitem[Pechenick et~al.(2015)Pechenick, Danforth, and
  Dodds]{10.1371/journal.pone.0137041}
Eitan~Adam Pechenick, Christopher~M. Danforth, and Peter~Sheridan Dodds.
\newblock Characterizing the {Google} books corpus: Strong limits to inferences
  of socio-cultural and linguistic evolution.
\newblock \emph{PLoS ONE}, 10\penalty0 (10):\penalty0 e0137041, 2015.

\bibitem[Piotrowski(2012)]{doi:10.2200/S00436ED1V01Y201207HLT017}
Michael Piotrowski.
\newblock Natural language processing for historical texts.
\newblock \emph{Synthesis Lectures on Human Language Technologies}, 2012.

\bibitem[Piper(2015)]{piper2015novel}
Andrew Piper.
\newblock Novel devotions: Conversional reading, computational modeling, and
  the modern novel.
\newblock \emph{New Literary History}, 46\penalty0 (1):\penalty0 63--98, 2015.

\bibitem[Piper(2017)]{piper2017think}
Andrew Piper.
\newblock Think small: On literary modeling.
\newblock \emph{PMLA}, 132\penalty0 (3):\penalty0 651--658, 2017.

\bibitem[Ribeiro et~al.(2016)Ribeiro, Singh, and
  Guestrin]{Ribeiro:2016:WIT:2939672.2939778}
Marco~Tulio Ribeiro, Sameer Singh, and Carlos Guestrin.
\newblock ``{{Why}} should {{I}} trust you?": Explaining the predictions of any
  classifier.
\newblock In \emph{Proceedings of KDD '16}, pages 1135--1144, 2016.

\bibitem[Salganik(2017)]{salganik2017bit}
Matthew~J. Salganik.
\newblock \emph{Bit by Bit: Social Research in the Digital Age}.
\newblock Princeton University Press, 2017.

\bibitem[Schofield and Mimno(2016)]{TACL868}
Alexandra Schofield and David Mimno.
\newblock Comparing apples to apple: The effects of stemmers on topic models.
\newblock \emph{Transactions of the Association for Computational Linguistics},
  4:\penalty0 287--300, 2016.

\bibitem[Sproat et~al.(2001)Sproat, Black, Chen, Kumar, Ostendorf, and
  Richards]{sproat2001}
Richard Sproat, Alan~W. Black, Stanley Chen, Shankar Kumar, Mari Ostendorf, and
  Christopher Richards.
\newblock Normalization of non-standard words.
\newblock \emph{Computer Speech \& Language}, 15\penalty0 (3):\penalty0 287 --
  333, 2001.

\bibitem[Srivastava et~al.(2018)Srivastava, Goldberg, Manian, and
  Potts]{doi:10.1287/mnsc.2016.2671}
Sameer~B. Srivastava, Amir Goldberg, V.~Govind Manian, and Christopher Potts.
\newblock Enculturation trajectories: Language, cultural adaptation, and
  individual outcomes in organizations.
\newblock \emph{Management Science}, 64\penalty0 (3):\penalty0 983--1476, 2018.

\bibitem[Tangherlini(2016)]{jaf2016tim}
Timothy~R. Tangherlini.
\newblock Big folklore: A special issue on computational folkloristics.
\newblock \emph{The Journal of American Folklore}, 129\penalty0 (511):\penalty0
  5--13, 2016.

\bibitem[Tausczik and Pennebaker(2010)]{liwc2010}
Yla~R. Tausczik and James~W. Pennebaker.
\newblock The psychological meaning of words: {LIWC} and computerized text
  analysis methods.
\newblock \emph{Journal of Language and Social Psychology}, 29\penalty0
  (1):\penalty0 24--54, 2010.

\bibitem[Tromble et~al.(2017)Tromble, Storz, and Stockmann]{tromble2017we}
Rebekah Tromble, Andreas Storz, and Daniela Stockmann.
\newblock We don't know what we don't know: When and how the use of {Twitter}'s
  public {APIs} biases scientific inference.
\newblock \emph{Available at SSRN: https://ssrn.com/abstract=3079927}, 2017.

\bibitem[Waseem and Hovy(2016)]{waseem-hovy:2016:N16-2}
Zeerak Waseem and Dirk Hovy.
\newblock Hateful symbols or hateful people? {Predictive} features for hate
  speech detection on {Twitter}.
\newblock In \emph{Proceedings of the NAACL Student Research Workshop}, pages
  88--93, San Diego, California, 2016.

\bibitem[Webb et~al.(1966)Webb, Campbell, Schwartz, and
  Sechrest]{webb1966unobtrusive}
Eugene~J. Webb, Donald~Thomas Campbell, Richard~D. Schwartz, and Lee Sechrest.
\newblock \emph{Unobtrusive Measures: Nonreactive Research in the Social
  Sciences}.
\newblock Chicago, IL: Rand McNally, 1966.

\bibitem[Williams et~al.(2017)Williams, Burnap, and
  Sloan]{doi:10.1177/0038038517708140}
Matthew~L. Williams, Pete Burnap, and Luke Sloan.
\newblock Towards an ethical framework for publishing {Twitter} data in social
  research: Taking into account users' views, online context and algorithmic
  estimation.
\newblock \emph{Sociology}, 51\penalty0 (6):\penalty0 1149--1168, 2017.

\bibitem[Zubiaga et~al.(2016{\natexlab{a}})Zubiaga, Kochkina, Liakata, Procter,
  and Lukasik]{zubiaga-EtAl:2016:COLING}
Arkaitz Zubiaga, Elena Kochkina, Maria Liakata, Rob Procter, and Michal
  Lukasik.
\newblock Stance classification in rumours as a sequential task exploiting the
  tree structure of social media conversations.
\newblock In \emph{Proceedings of COLING 2016, the 26th International
  Conference on Computational Linguistics: Technical Papers}, pages 2438--2448,
  Osaka, Japan, 2016{\natexlab{a}}.

\bibitem[Zubiaga et~al.(2016{\natexlab{b}})Zubiaga, Liakata, Procter, Hoi, and
  Tolmie]{zubiaga2016analysing}
Arkaitz Zubiaga, Maria Liakata, Rob Procter, Geraldine Wong~Sak Hoi, and Peter
  Tolmie.
\newblock Analysing how people orient to and spread rumours in social media by
  looking at conversational threads.
\newblock \emph{PloS one}, 11\penalty0 (3):\penalty0 e0150989,
  2016{\natexlab{b}}.

\end{thebibliography}
\end{document}